%% file: main_iaai.tex
\documentclass[letterpaper]{article} 
\usepackage{aaai2026}  
\usepackage{times}  
\usepackage{helvet}  
\usepackage{enumitem}
\usepackage{courier}  
\usepackage[hyphens]{url}  
\usepackage{graphicx} 
\usepackage{natbib}  
\usepackage{caption} 
\usepackage{algorithm}

\usepackage{newfloat}
\usepackage{listings}
\DeclareCaptionStyle{ruled}{labelfont=normalfont,labelsep=colon,strut=off} 
\floatstyle{ruled}
\newfloat{listing}{tb}{lst}{}
\floatname{listing}{Listing}
\title{Prediction of Hospital Associated Infections During Continuous Hospital Stays}
\author{
    Rituparna Datta\textsuperscript{\rm 2},
    Methun Kamruzzaman\textsuperscript{\rm 6},
    Eili Y. Klein\textsuperscript{\rm 3},
    Gregory R. Madden\textsuperscript{\rm 4},
    Xinwei Deng\textsuperscript{\rm 5},
    Anil Vullikanti\textsuperscript{\rm 1,2},
    Parantapa Bhattacharya\textsuperscript{\rm 1}
}
\affiliations{
    \textsuperscript{\rm 1}Biocomplexity Institute \& Initiative, University of Virginia\\
    \textsuperscript{\rm 2}Department of Computer Science, University of Virginia\\
    \textsuperscript{\rm 3}Johns Hopkins University School of Medicine and Johns Hopkins Center for Health Security\\
    \textsuperscript{\rm 4}Division of Infectious Diseases \& International Health, University of Virginia Health System\\
    \textsuperscript{\rm 5}Department of Statistics, Virginia Tech,
    \textsuperscript{\rm 6}Systems Biology, Sandia National Laboratory
}

\newcommand{\tool}{GenHAI}
\newcommand{\probname}{\emph{Hospital Test Sequence Modeling}}
\newcommand{\prob}{HTSM}

\usepackage{bibentry}
\input{preamble}
\begin{document}

\maketitle
\input{abstract2}

\input{intro}

\input{related-short}
\input{problem}

\input{method_short}
\input{experiments}
\input{case-study2}

\input{conclu}

\bibliography{aaai2026}
\clearpage
\appendix
\input{related}

\input{indiv-models}
\input{appendix-method}
\input{limitation}
\end{document}

%% file: preamble.tex
\usepackage{algpseudocode}
\usepackage{mathtools}
\usepackage{caption}
\usepackage{subcaption}
\usepackage{booktabs}
\usepackage{bm}
\usepackage{amsfonts}

\DeclareMathOperator{\Normal}{\mathcal{N}}
\DeclareMathOperator{\Bernoulli}{\mathcal{B}}
\DeclareMathOperator{\NegBinom}{\mathcal{NB}}

\DeclareMathOperator{\Multinomial}{{\mathcal{M}}}

\newcommand{\balpha}{\boldsymbol{\alpha}}
\newcommand{\bbeta}{\boldsymbol{\beta}}
\newcommand{\bmu}{\boldsymbol{\mu}}
\newcommand{\bSigma}{\boldsymbol{\Sigma}}

\newcommand{\btheta}{\boldsymbol{\theta}}

\newcommand{\bNormal}{\boldsymbol{\mathcal{N}}}

\newcommand{\cD}{\mathcal{D}}

\newcommand{\bw}{\mathbf{w}}
\newcommand{\bx}{\mathbf{x}}

\newcommand{\bD}{\mathbf{D}}

\newcommand{\bbD}{\mathbb{D}}
\newcommand{\bbQ}{\mathbb{Q}}
\newcommand{\bbI}{\mathbb{I}}
\newcommand{\bbE}{\mathbb{E}}

\newcommand{\Th}[1]{\ensuremath{#1^{\text{th}}}}

\newcommand{\Tab}{\text{ab}}
\newcommand{\Ticu}{\text{icu}}
\newcommand{\Tdia}{\text{dia}}
\newcommand{\Tcont}{\text{cont}}

\newcommand{\UVADataset}{UVA Dataset}

%% file: abstract2.tex
\begin{abstract}
The US Centers for Disease Control and Prevention (CDC), in 2019,
designated Methicillin-resistant \emph{Staphylococcus aureus} (MRSA)
as a serious antimicrobial resistance threat.
The risk of acquiring MRSA and suffering life-threatening consequences due to it
remains especially high for hospitalized patients due to a unique combination of factors, including:
comorbid conditions, immunosuppression, and antibiotic use,
and risk of contact with contaminated hospital workers and equipment.
In this paper, we present a novel generative probabilistic model, \tool{},
for modeling sequences of MRSA test results outcomes
for patients during a single hospitalization.
This model can be used to answer many important questions
from the perspectives of hospital administrators
for mitigating the risk of MRSA infections.
Our model is based on the probabilistic programming paradigm,
and can be used to approximately answer a variety of
predictive, causal, and counterfactual questions.
We demonstrate the efficacy of our model
by comparing it against discriminative and generative machine learning models
using two real world datasets.
\end{abstract}

%% file: intro.tex
\section{Introduction}\label{sec:intro}

\begin{figure*}
    \centering
\includegraphics[width=0.72\linewidth]{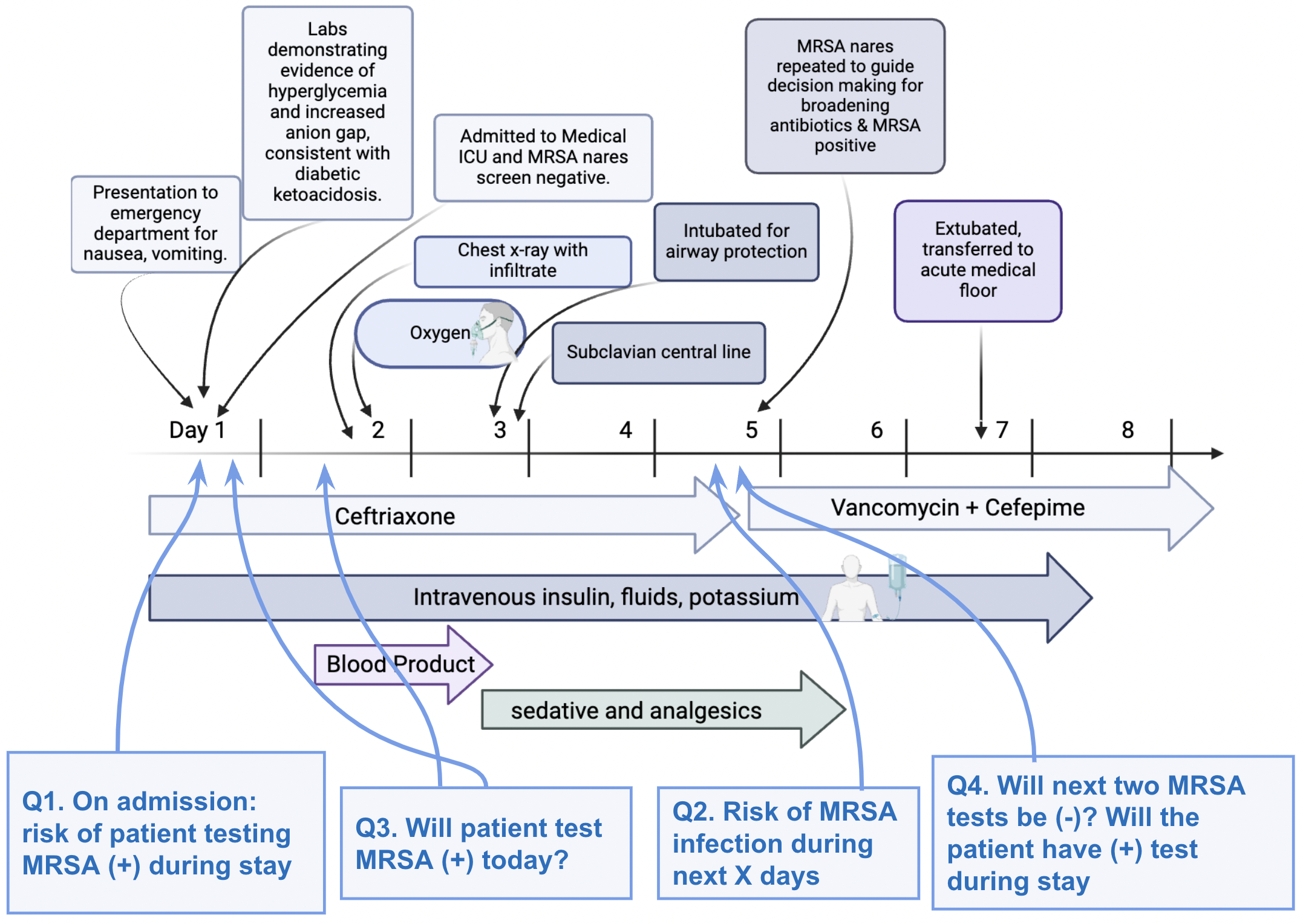}
\caption{
\small
Examples of different questions related to infection risk for a patient during their hospital stay.
Q1 arises on patient admission, while Q2, Q3, and Q4 are examples of questions clinicians ask during the patient's stay in the hospital.
}
\label{fig:patient-workflow}
\end{figure*}

Complex workflows are associated with patient care in hospitals, even for seemingly routine conditions.
Figure~\ref{fig:patient-workflow} shows the initial part of the workflow associated with a patient who comes into the Emergency Department.
The patient is put on antibiotics and multiple procedures are done as part of diagnoses, and then moved to an ICU.
There is often a significant risk of Healthcare-associated infections (HAIs), such as Methicillin-resistant \emph{Staphylococcus aureus} (MRSA) during the patient's stay.
HAIs lead to longer hospital stays, increased mortality~\cite{weiner:iche2020,10.1001/jamainternmed.2013.10423}, and billions of dollars in increased healthcare costs \cite{10.1001/jamainternmed.2013.9763}.
While regular antibiotics work for bacterial infections, more aggressive antibiotics are used for a patient with an HAI.
Further, HAIs can be transmitted within the hospital, and patients with HAIs are isolated~\cite{cui2024modeling}. 
Early detection and effective prediction of the risk of acquisition and severe outcomes from HAIs infections can help in judicious and targeted antibiotic administration and implementation of better isolation precautions. Therefore, clinicians constantly ask about the risk of their patients getting MRSA or other HAIs during their hospital stay; examples of such questions are indicated as  Q1-Q4 in Figure~\ref{fig:patient-workflow}.  Some of these questions might be asked multiple times during the patient's stay, as clinicians pick the best treatments.
This has motivated a lot of work on ML methods for risk prediction type problems using Electronic Health Record (EHR) data for MRSA (the focus here) and other HAIs, e.g.,~\cite{Hartvigsen2018EarlyPO,oh:iche18,Monsalve15,kamruzzaman2024improving,chang:plosone11,dubberke:iche11,na:pone15,modayil:gastroenterology11,fan2022random,liu2019influencing,huynh2022domain,ambavane20232814}. These methods, which are generally discriminative, have shown that specific types of clinical predictions, e.g., risk of HAI infection within $d$ days, can be done reasonably well, e.g,~\cite{Monsalve15,kamruzzaman2024improving}.

Despite the success of EHR-based ML methods for some of HAI related questions that arise, their use in actual clinical practice is somewhat less; this is a significant contrast to other areas of clinical informatics,  where ML has been used effectively, e.g.,~\cite{simpao2014review, nordo2019use}. While there are multiple reasons, an important issue is that the complex workflows in clinical practice related to HAIs require a decision-theoretic setting, in which risk prediction needs to be done in a variety of scenarios. 
As mentioned earlier, Q1-Q4 in Figure~\ref{fig:patient-workflow} are examples of questions related to HAIs which clinicians ask at different stages for their patients. 
Some of these questions (e.g., Q1, Q4) are counterfactual in nature, and arise when clinicians evaluate treatments (including antibiotic choice) and precautions.
For instance, a clinician would use Q1, Q2 and Q3 to decide whether to order a an MRSA test, and if antibiotic treatments should be started or continued---these are key elements of antibiotic stewardship plans~\cite{cdc-antibiotic, cosgrove2023antibiotic}.
Q4 arises when clinicians are deciding whether to keep a patient under contact precautions to reduce risk to others, or remove them, e.g.,~\cite{cui2024modeling, marshall2013active, siegel20072007}.


Prior discriminative ML methods, e.g.,~\cite{Hartvigsen2018EarlyPO}, are developed and trained for a specific form of risk prediction, and cannot be used directly for such counter-factual settings, with robust uncertainty estimates. This could be done by learning a large collection of models, but these become very hard to interpret, and ensure consistency, and there is usually limited data for learning multiple models, especially in clinical settings. Additionally, uncertainty estimation and statistical calibration become challenging with such methods. This is critical in clinical settings, where decisions must weigh not just predicted risk but also the confidence and reliability of those predictions.

Generative models using probabilistic programming~\cite{probprog-gordon14,probprog-book-meent21} provide a natural approach to handle the limitations of discriminative methods, especially for healthcare applications~\cite{chen2021probabilistic,pmlr-v149-urteaga21a,Makar_Guttag_Wiens_2018}. They can provide very interpretable methods for modeling complex decision processes, while easily providing uncertainty estimates and statistical calibration. Generative models have supported diverse patient-level predictions, including menstrual cycle length~\cite{pmlr-v149-urteaga21a} and probability of infection in a networked model~\cite{Makar_Guttag_Wiens_2018}. However, the problems considered in prior work are much simpler clinical questions; they do not contain conditionals and loops, which arise naturally in clinical practice.

\noindent
{\textbf{Contributions}} (many details are presented in the Appendix)\\ 
(1) We present a novel generative sequence modeling problem that we call the \probname{} (\prob{}) problem. 
This problem is designed based on clinical problems arising from management of HAIs at the University of Virginia (UVA) and Johns Hopkins University (JHU) hospitals, which are large hospitals serving two very diverse populations.\\
(2) We present a novel modular interpretable probabilistic program, \tool{} (\textbf{Gen}erative Model for \textbf{H}ospital-\textbf{A}cquired \textbf{I}nfections), for modeling the generative process of MRSA test results. 
\tool{} can then be used to answer many important questions using probabilistic queries. 
The probabilistic program incorporates domain knowledge from clinical experts at UVA and JHU (Section~\ref{sec:method}); our results show that incorporating this domain knowledge (e.g., number of days on antibiotics, and ICU stays) helps in significantly improving the performance of \tool{}, and presents a template for other kinds of clinical questions.\\ 
(3) We evaluate \tool{} on EHR datasets from UVA and MIMIC-III~\cite{johnson2016mimic}, and demonstrate its goodness of fit.
We also compare it with state-of-the-art deep neural network-based generative sequence prediction models.
\tool{} generally has the best or close to the best performance in all metrics. 
Further, our approach naturally gives uncertainty estimates, and is easy to calibrate and interpret.\\ 
(4) We consider four case studies, each considering specific types of questions arising from clinical practice at the two hospitals, that can be answered using probabilistic queries and our probabilistic program.
\tool{} gives clinically interpretable patient level estimates, which are not possible using standard supervised ML methods.



%% file: related-short.tex
\section{Related Works}\label{sec:related-short}

We summarize the main threads of related work here; additional discussion is presented in Section~\ref{sec:related} in the Appendix.

Prior work has developed prediction tools for MRSA and other HAIs using EHR data~\cite{Hartvigsen2018EarlyPO,oh:iche18,Monsalve15,kamruzzaman2024improving,chang:plosone11,dubberke:iche11,na:pone15,modayil:gastroenterology11,fan2022random,liu2019influencing,huynh2022domain,ambavane20232814}. While simple ML with careful feature engineering can capture some risks~\cite{Monsalve15,kamruzzaman2024improving,fan2022random}, such models are limited to narrow classification tasks and cannot support the broad queries clinicians face.  
Our work is also related to sequence modeling for clinical prediction, e.g., ~\cite{zhang2023crossformer, Ekambaram_2023, mamba,mamba2, nguyen2021clinical}, some of which are used as baselines. 




Probabilistic machine learning and generative models have been proposed for many clinical tasks, e.g.,~\cite{chen2021probabilistic,pmlr-v149-urteaga21a,Makar_Guttag_Wiens_2018}. \cite{pmlr-v149-urteaga21a} develop a simple but flexible generative model for the prediction of menstrual cycle length using a hierarchical, Generalized Poisson-based generative model, which explicitly incorporates individual behaviors, and produces calibrated individual predictions.
\cite{Makar_Guttag_Wiens_2018} show that a generative probabilistic model performs very well for the problem of predicting the activation of an individual in a networked epidemic process. 

%% file: problem.tex
\section{Background}\label{sec:problem}



As detailed in Section~\ref{sec:dataset}, EHRs capture comprehensive information about each patient’s hospitalization, which we leverage in this work. A central element for our study is the MRSA testing data, consisting of two types: NARE tests (PCR-based) and culture tests. NARE tests are explicitly performed to detect MRSA, while culture tests may be done for a variety of clinical purposes and are not always intended as MRSA screenings. However, if MRSA is identified in a culture test, the result is documented in the patient’s healthcare record. In our datasets, all NARE test results are available for each hospitalization record, whereas culture tests are recorded only when the result is MRSA-positive. We found this to be a common data issue when gathering MRSA test result data, based on our discussions with clinicians at UVA and JHU, as well as in the MIMIC-III dataset.

\begin{table}[h!]
\centering
\caption{Table of notation}
\resizebox{\linewidth}{!}{%
\begin{tabular}{p{2cm}|p{6.5cm}}
\hline
\textbf{Symbol} & \textbf{Description} \\
\hline
$\textbf{D}$ & Dataset of hospitalization records \\
$n$ & Number of hospitalization records \\
$p_k$ & Patient corresponding to the $k$-th hospitalization record \\
$m_k$ & Number of MRSA tests in the $k$-th record \\
$r_{k,i}$ & Outcome of the $i$-th MRSA test for patient $p_k$ \\
$t_{k,i}$ & Test type of the $i$-th MRSA test (NARE or culture) \\
$d_{k,i}$ & Time interval between tests $t_{k,i}$ and $t_{k,i+1}$ \\
$\alpha_k$ & Admission-time features of patient $p_k$ \\
$\beta_{k,i}$ & Features of patient $p_k$ at the time of test $t_{k,i}$ \\
$\cD_{*}$ & Probabilistic sub-programs (e.g., $\cD_{t_i}, \cD_{r_i}, \cD_{d_i|+}$) \\
$\cD_{\beta_{[ab]}}$,$\cD_{\beta_{[icu]}}$, $\cD_{\beta_{[dia]}}$ & Distributions for days on antibiotics (30d), in ICU (7d), dialysis (7d) \\
$\cD_{d_i|{+,-}}$ & Inter-test delay distribution when previous result is positive/ negative \\
$\cD_{\text{cont}}$ & Bernoulli continuation distribution (decides if sequence continues) \\
\hline
\end{tabular}}
\label{tab:notation}
\end{table}

Here we formally describe the assumptions made about the patient healthcare care records and the overall modeling problem;
the notation we use is summarized in Table~\ref{tab:notation}.
We assume that the dataset $\bD$ consists of $n$ hospitalization records. Let $p_k$ ($k\in [1\ldots n]$) be the patient corresponding to the \Th{k} hospitalization record. Additionally, let $\{r_{k,i}\}$ and $\{t_{k,i}\}$ ($i \in [1\ldots m_k]$) be the sequence of observable test outcomes and the corresponding test type (NARE test or culture test) in the \Th{k} hospitalization record. Here $m_k$ is the number of MRSA test results in the \Th{k} hospitalization record. Further, let $d_{k,i}$ be the time duration in between tests $t_{k,i}$ and $t_{k,i+1}$, where \mbox{$i \in [1\ldots {m_k-1}]$}. Also, let $\balpha_k$ be the vector representing the set of features of the patient $p_k$ that are observable at admission time, and let $\bbeta_{k,i}$ be a vector representing the set of observable features of the patient $p_k$ at the time when the test $t_{k,i}$ was done. We assume that each hospital sequence is independent of every other sequence.

\noindent
\textbf{Problem Statement.}
The objective of the \probname{} (\prob{}) is to learn the joint conditional probability distribution:
\begin{equation}
\small
  P(\bD) = \prod_{k=1}^n P(\{r_{k,i}\}, \{t_{k,i}\}, \{d_{k,i}\}, \{\bbeta_{k,i}\} \mid \balpha_k)
\end{equation}

Note that we do not model $\balpha_k$
which represents the patient's characteristics available at the time of admission ---
such as gender, age, past MRSA test results, etc. ---
but consider it as an input to our model.
Given the distribution $P(\bD)$
all of the questions posed in Section~\ref{sec:intro}
can then be answered using probabilistic queries using the model.

%% file: method_short.tex
\section{Modeling Test Result Sequences}\label{sec:method}

We take a generative modeling approach for solving the \prob{} problem.
In particular, we describe the generative process
using a modular probabilistic program.
A probabilistic program is generative model
that takes the form of an imperative or functional program with support for sampling random variables
and conditioning the values of those variables to observations~\cite{probprog-gordon14}.
Probabilistic programs are Bayesian models.
They are inherently more interpretable
and allow for easy incorporation of domain knowledge from experts
compared to deep neural network-based models.
They are, however, generally harder to use due to the computational complexity of training them.

To alleviate the issue of computational complexity, the probabilistic program used here is a modular one, composed of a number of smaller probabilistic programs that can be trained independently. This is made possible by ensuring that the top-level probabilistic program doesn't contain any latent variables, allowing parallel training and making the model easier to debug and interpret.

\paragraph{Overview.}
We use a probabilistic program to model the generative process for hospital test sequences.
The program uses a number of sub-programs: $\cD_{\bbeta_i[\Tab]}$, $\cD_{\bbeta_i[\Ticu]}$, $\cD_{\bbeta_i[\Tdia]}$, $\cD_{t_1}$, $\cD_{r_i}$, $\cD_{\Tcont}$, $\cD_{d_i|-}$, $\cD_{d_i|+}$; these are denoted by $\cD_{*}$ for convenience, and are described below.
Each sub-program takes the form of a
Bayesian model with a Generalized Linear Model (GLM) like design.

Consider the case of a generic sub-program $\cD_Y$
which is used to sample the random variable $Y$
given the input parameters $\bx$.
Subprogram $\cD_Y$ can be defined
using the output distribution $\bbD_Y$,
parameters $\btheta_Y$, link function $\ell_Y$,
and a distribution on the model parameters.
The random variable $Y$ is sampled from $\cD_Y$ as follows:
\noindent
{\small
\[
  Y \sim \bbD_Y(\theta_i, \theta_d);\hspace{1em}  
  \theta_d = \ell_Y(\bw^T \cdot \bx + c); \hspace{1em}
  \theta_i, \bw, c \sim \bbQ(\btheta_Y)
\] 
}
Here $\theta_i$ and $\theta_d$ represent the input independent
and input dependent parameters of $\bbD_Y$,
$(\bw, c)$ are the linear model parameters,
and $\bbQ(\btheta_Y)$ represents the posterior distribution
of the model parameters $(\theta_i, \bw, c)$, conditioned on data.

Each of the individual sub-programs $\cD_{*}$ is trained using \textit{Stochastic Variational Inference (SVI)} by maximizing the \textit{Evidence Lower Bound (ELBO)} objective~\cite{elbo,bbvi} We use Multivariate Normal distributions
as variational approximation of the posterior distributions $\bbQ_{*}$.

\subsubsection{Algorithm \tool{} (informal).}\label{ssection:simulation}
The generative process simulates a patient’s MRSA test sequence during hospitalization. The simulation of the program starts by sampling admission-time features ($\alpha$) and initial test-time features ($\beta_1$). At each step, it 
\\\textbf{(i)} samples test type, $t_i \sim \cD_{t_i}(\cdot)$ and 
result $r_i$ (with $r_i=1$ deterministically if $t_i$ is culture, otherwise $r_i \sim \cD_{r_i}(\cdot)$)\\
\textbf{(ii) }draws inter-test delay ($d_i$) using a Log-Normal mixture if the previous result ($r_{i-1}$) was negative, or a single Log-Normal otherwise\\
\textbf{(iii)} updates test-time covariates ($\beta_i$) via $\cD_{\beta[ab]},  \cD_{\beta[icu]},  \\\cD_{\beta[dia]}$, modeling recent antibiotics (30d), ICU stays (7d), and dialysis status. The first two are parameterized as truncated Negative Binomial distributions to account for overdispersed count data with hard upper bounds, while $\cD_{\beta[dia]}$ is modeled as a Bernoulli random variable. 
\\
\textbf{(iv)} applies a Bernoulli continuation rule ($\cD_{\text{cont}}$) and repeats until termination, yielding a synthetic test sequence conditioned on $\alpha$. 

The complete program (Algorithm~\ref{alg:full-sim}, Appendix) expands this workflow into modular sub-programs $\cD_{*}$ with explicit distributional forms and inference details.


\begin{figure}
  \centering
  \begin{subfigure}[b]{0.48\linewidth}
    \centering
    \includegraphics[width=\linewidth]{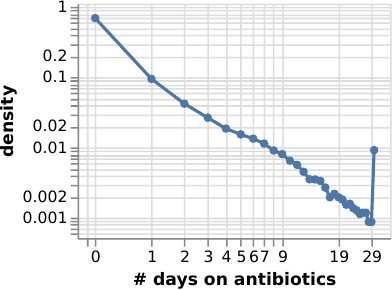}
    \caption{Distribution of the number of days on antibiotics, within last 30 days.\label{fig:ab-days-hist}}
  \end{subfigure}
  \hfill
  \begin{subfigure}[b]{0.48\linewidth}
    \centering
    \includegraphics[width=\linewidth]{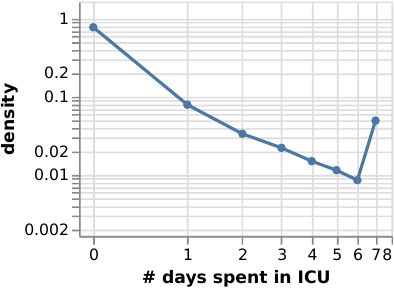}
    \caption{Distribution of the number of days spent in the Intensive Care Unit, within the past 7 days.\label{fig:icu-days-hist}}
  \end{subfigure}
  \caption{Distribution of features observable at test time.\label{fig:test-time-features}}
\end{figure}

\begin{figure}
  \centering
  \includegraphics[width=1\linewidth]{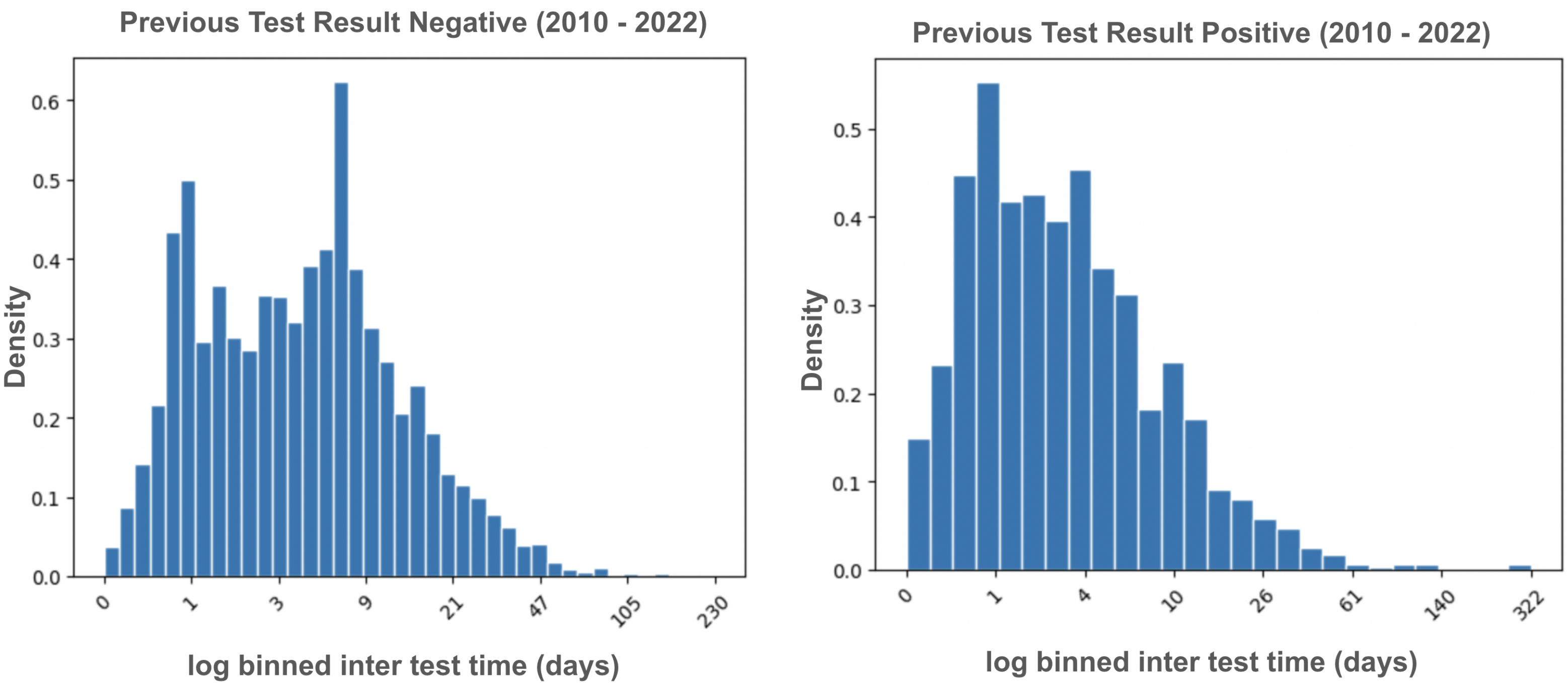}
  \caption{Distribution of inter test time}\label{fig:inter-test-time-hist}
  \vspace{-1em}
\end{figure}
Our model simulates a patient’s sequence of MRSA tests during hospitalization. It initializes patient features such as days on antibiotics, ICU days, and dialysis status (Figures~\ref{fig:ab-days-hist}, \ref{fig:icu-days-hist}). At each step, the model samples the test type (culture or NARE), result, and inter-test delay, which depends on the previous result. Negative tests follow a mixture of Log-Normal delays, while positives use a single Log-Normal (Figure~\ref{fig:inter-test-time-hist}). The sequence ends via a Bernoulli stopping rule; additional details in Appendix~\ref{appendix:alg1-description}.

\paragraph{Implementation.}
The overall probabilistic program is implemented in Python.
The individual sub-programs $\cD_{*}$ were created
using the Pyro probabilistic programming language~\cite{pyro}
which internally uses the PyTorch~\cite{pytorch} deep neural network library
for auto-differentiation and speeding up computations on hardware accelerators.

%% file: experiments.tex
\section{Experiments}\label{sec:experiments}

\input{dataset}

\subsection{Baselines}
We compare our probabilistic program (\tool{}) with recent state-of-the-art deep learning models for sequence prediction, including Crossformer~\cite{zhang2023crossformer}, TSMixer~\cite{Ekambaram_2023}, Mamba~\cite{mamba, mamba2}, and TPAMTL~\cite{nguyen2021clinical} (Details in Appendix). These baselines were selected for their relevance, availability, and performance on time-series tasks. Table~\ref{tab:clf_metrics} compares \tool{} with baseline models, showing that while standard sequence models achieve reasonable accuracy, \tool{} provides competitive performance with better calibration and interpretability.

\subsection{Training}
All models were trained on machines equipped with
four Nvidia Tesla V100 GPUs,
two 20-core Intel Xeon Gold 6230 CPUs,
and 380 GB of RAM. Each of the sub-programs used in Algorithm~\ref{alg:full-sim}
were trained using Stochastic Variational Inference (SVI)~\cite{elbo,bbvi}
using the Pyro probabilistic programming language~\cite{pyro}.
Details about the hyperparameters used for training
as well as the prior distributions used in the Bayesian models
are provided in Appendix.

\subsection{Results}
We evaluate the negative log-likelihood (NLL) and perplexity of each \tool{} sub-program ($\cD_{*}$) on the UVA and MIMIC-III datasets, using an 80:20 train–test split. Across both datasets, the sub-programs achieve consistently low NLL and perplexity values, ranging from 0.135–2.33 (UVA) and 0.018–2.006 (MIMIC-III) in NLL, and 1.019–10.283 (UVA) and 1.027–7.433 (MIMIC-III) in perplexity, indicating that the generative components capture realistic temporal dynamics of patient MRSA testing sequences (Table \ref{tab:gen_metrics}, Appendix).
This validation confirms that the model components are statistically calibrated and provide reliable likelihood estimates for observed sequences. Such calibration is essential, since the overall framework depends on these sub-programs to support downstream predictive and counterfactual queries. Unlike discriminative models that output only point estimates, \tool{} produces full posterior distributions, enabling uncertainty quantification. This ability to report both predicted risk and confidence makes \tool{} well-suited for clinical decision-making, where uncertainty directly influences isolation, antibiotic use, and retesting. Here, the goal is not classification accuracy but generative fit, ensuring that probabilistic reasoning rests on realistic patient-level distributions rather than heuristic approximations.

\input{clf_metrics_table}

Table~\ref{tab:clf_metrics} compares the performance of the four deep neural network-based baselines for the NARE test result prediction task. Notably, TPAMTL~\cite{nguyen2021clinical} is the only model specifically designed for healthcare EHR data and evaluated on MIMIC-III using a probabilistic multi-task learning framework.
Despite its simplicity, \tool{}-$\cD_{r_i}$ achieves about $6.6\%$ higher accuracy than the next best model while maintaining competitive recall and F1.



%% file: dataset.tex
\subsection{Datasets}\label{sec:dataset}


For this study, we have used two datasets
for modeling MRSA test result sequences:
(1) \UVADataset{}, and
(2) the MIMIC-III v1.4 dataset~\cite{johnson2016mimic}.
An overview of the dataset statistics and feature construction details is provided in Section~\ref{sec:feature_construct}, and Table~\ref{table:patient_numbers} (Appendix).


\paragraph{\UVADataset{}.}
The \UVADataset{} is derived from Electronic Health Records at the UVA, and is Institutional Review Board (IRB)-restricted for authorized research use. It contains over 27K patients, 37K hospitalizations, and spans 2012–2022. Each hospitalization includes:\\
\textbf{1. On-admission data.} -- demographics (age, gender, race, ethnicity) and visit details (reason, type, source), used for admission-time features $\balpha$. \textbf{2. Clinical event data.} -- procedures (ICU stays), medications (e.g., antibiotics), comorbidities, and lab tests (including MRSA), used for test-time features $\bbeta$, test types $t_i$, and results $r_i$. \textbf{3 MRSA test data} -- clinical cultures (MRSA-positive only) and surveillance NARE tests (both positive and negative).

\paragraph{The MIMIC-III v1.4 dataset.}
The MIMIC-III dataset contains deidentified hospitalization records from over 40,000 patients at Beth Israel Deaconess Medical Center (2001–2012), including demographics, vitals, and labs, and is publicly available for clinical and epidemiological research~\cite{johnson2016mimic}.

%% file: clf_metrics_table.tex
\begin{table}
\resizebox{\linewidth}{!}{%
\begin{tabular}{l|r|r|r|r|r|r}
\toprule
  {\small Model} & {\small Accuracy} & {\small Precision} & {\small Recall} & {\small F1 Score} & {\small AUROC} & {\small AUPRC}\\
\midrule
  Mamba & 0.749 & 0.876 & 0.834 &  0.855 & 0.820 & 0.727 \\
  Crossformer & 0.809 & 0.650 &  0.639 & 0.644  & 0.825 & 0.606\\
  TSMixer & 0.758 & 0.817 &  0.758 & 0.786& 0.886 & 0.790 \\
  TPAMTL & 0.873 & 0.697 &  0.297  & 0.416 & 0.689 & 0.258\\
  \midrule
   \tool{}-$\cD_{r_i}$& 0.931 & 0.690 & 0.745 &  0.714& 0.858 & 0.635\\
\bottomrule
\end{tabular}}
  \caption{Comparing performance of baselines with $\cD_{r_i}$
  for MRSA test result prediction task for NARE tests.}\label{tab:clf_metrics}
\end{table}



%% file: case-study2.tex
\section{Case Studies}\label{sec:case-study}

In this section, we present answers to the real-world questions posed in Section~\ref{sec:intro}.
Although \tool{}’s simulation (Section~\ref{ssection:simulation}) can directly answer the probabilistic queries via rejection sampling, this approach is highly inefficient. Thus, for answering questions in case studies 2--4, we create variants of \tool{}.
The variants, however, utilize the same pre-trained sub-programs (i.e., $\cD_{*}$).
Details of the variant algorithms are described in detail in the Appendix.

\noindent
\paragraph{Case Study 1 (Admission Risk).}\label{ssec:case-study-1} \textit{A 70-year-old patient is admitted from a long-term care facility. The infection-control nurse wants to know: Should we place this patient in isolation on admission?}

\noindent
\textbf{A1:}
Let $\bbI_{A}(\{r_{i}\})$ be the indicator function that returns $1$ if any of the results are positive.
Then the answer to question Q1 can be obtained as:
{\small
\[
  P_A = \int \bbI_A (\{r_{i}\}) \approx \bbE [ \bbI_A (\{r_{i}\}) ]
\]
Using \tool{}, we can generate sample sequences $\{ r_i \}$ which can be used in the above equation to estimate the relevant likelihood. The system shows a higher risk for patients transferred from facilities compared to community admissions (Figure \ref{fig:cs1}), supporting proactive isolation and early testing for this patient group.

\begin{figure*}[t]
  \centering
  \captionsetup[sub]{font=footnotesize,labelfont=bf,justification=centering,aboveskip=4pt,belowskip=0pt}
\captionsetup{aboveskip=4pt,belowskip=0pt}

  \begin{subfigure}[t]{0.5\textwidth}
    \centering
    \includegraphics[width=0.85\linewidth,height = 4cm]{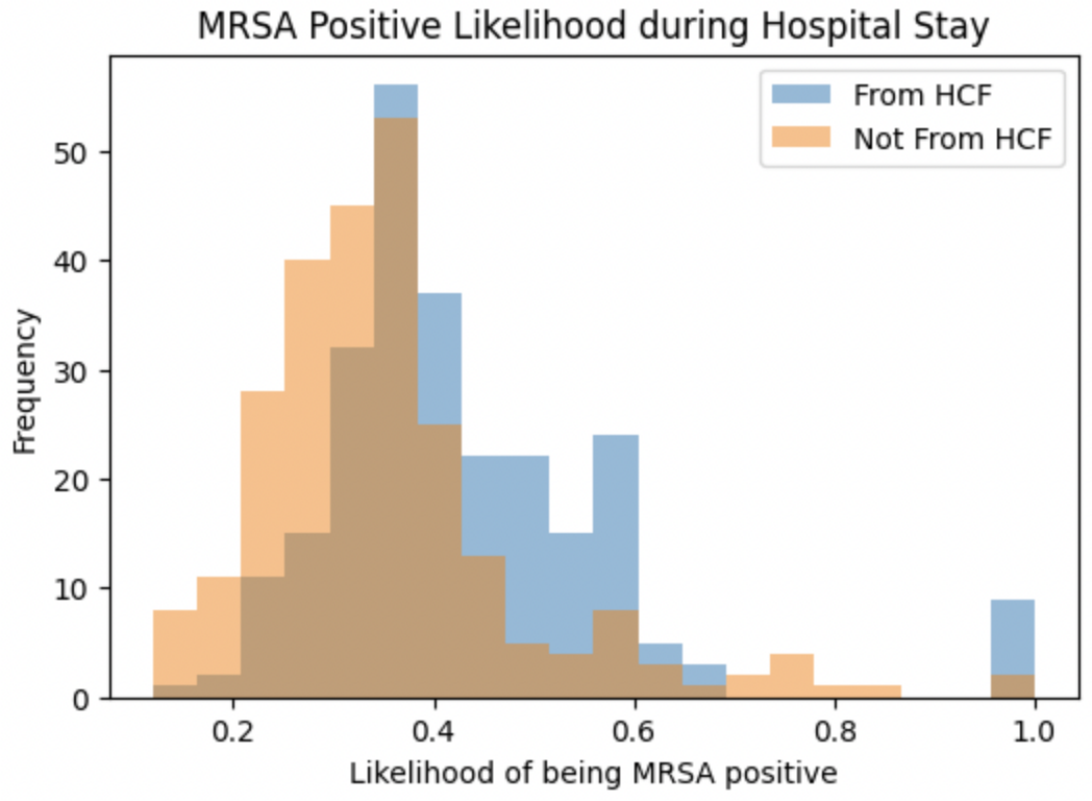}
    \caption{Distribution of MRSA-positive risk during stay by admission source(From HCF vs. Not From HCF)}
    \label{fig:cs1}
  \end{subfigure}\hfill
  \begin{subfigure}[t]{0.5\textwidth}
    \centering
    \includegraphics[width=0.85\linewidth,height = 4cm]{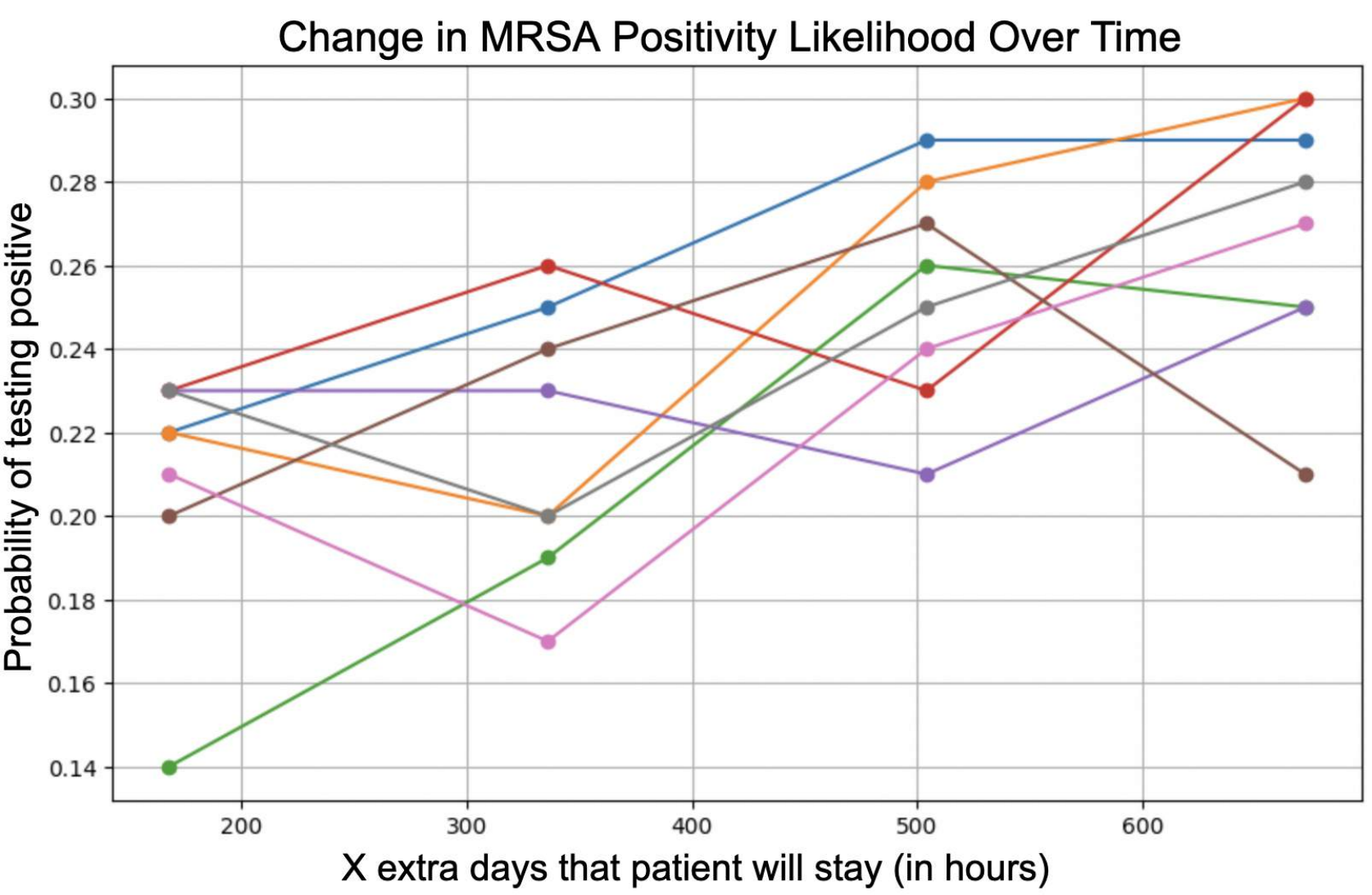}
    \caption{Change in MRSA-positive risk with the increasing number of additional days spent in the hospital (selected patients)}
    \label{fig:cs2}
  \end{subfigure}

  \begin{subfigure}[t]{0.5\textwidth}
    \centering
    \includegraphics[width=0.85\linewidth, height = 4cm]{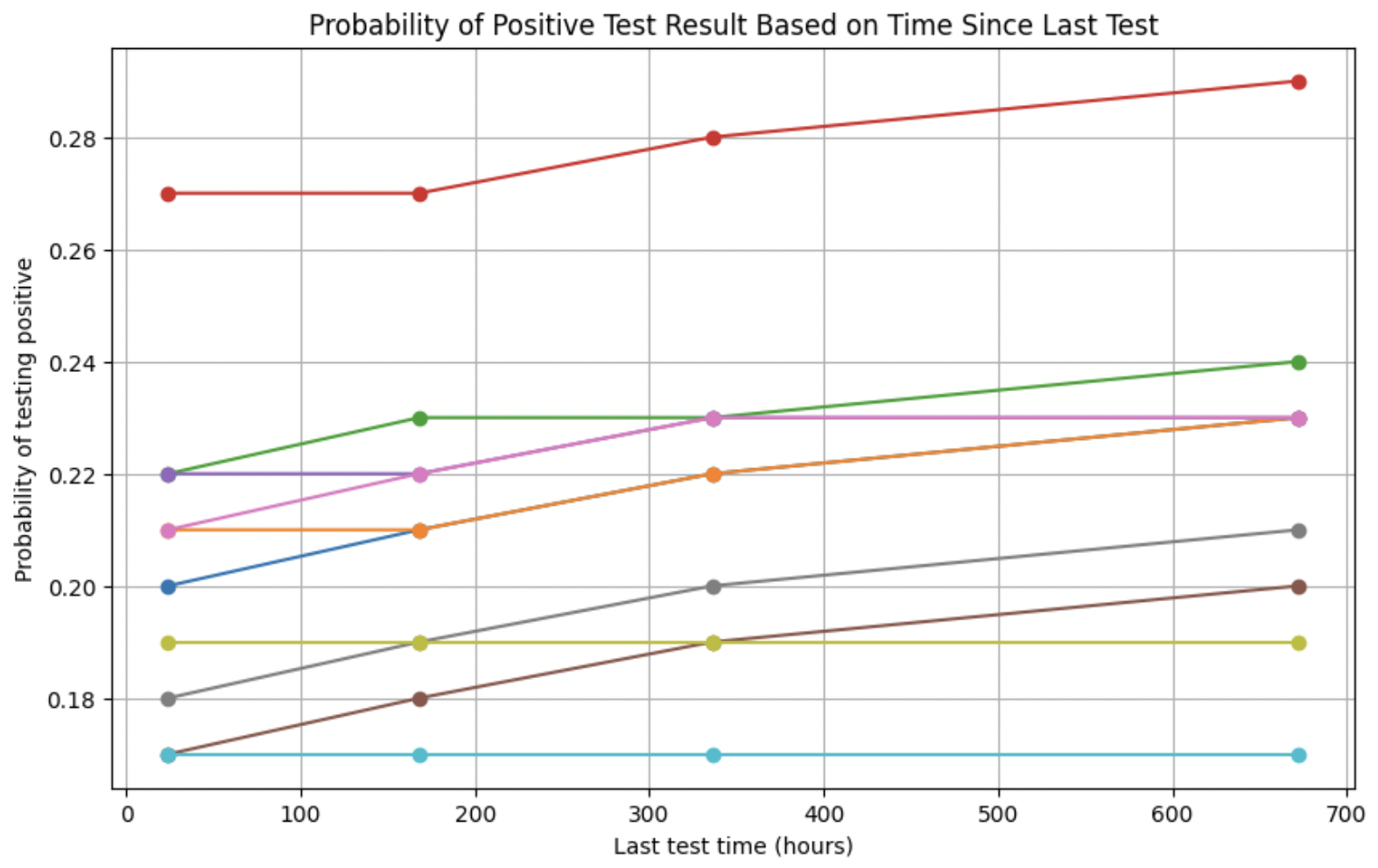}
    \caption{Change in MRSA-positive risk in the subsequent test with increasing delay between
previous and subsequent tests.}
    \label{fig:cs3}
  \end{subfigure}\hfill
  \begin{subfigure}[t]{0.5\textwidth}
    \centering
    \includegraphics[width=0.85\linewidth,height = 4cm]{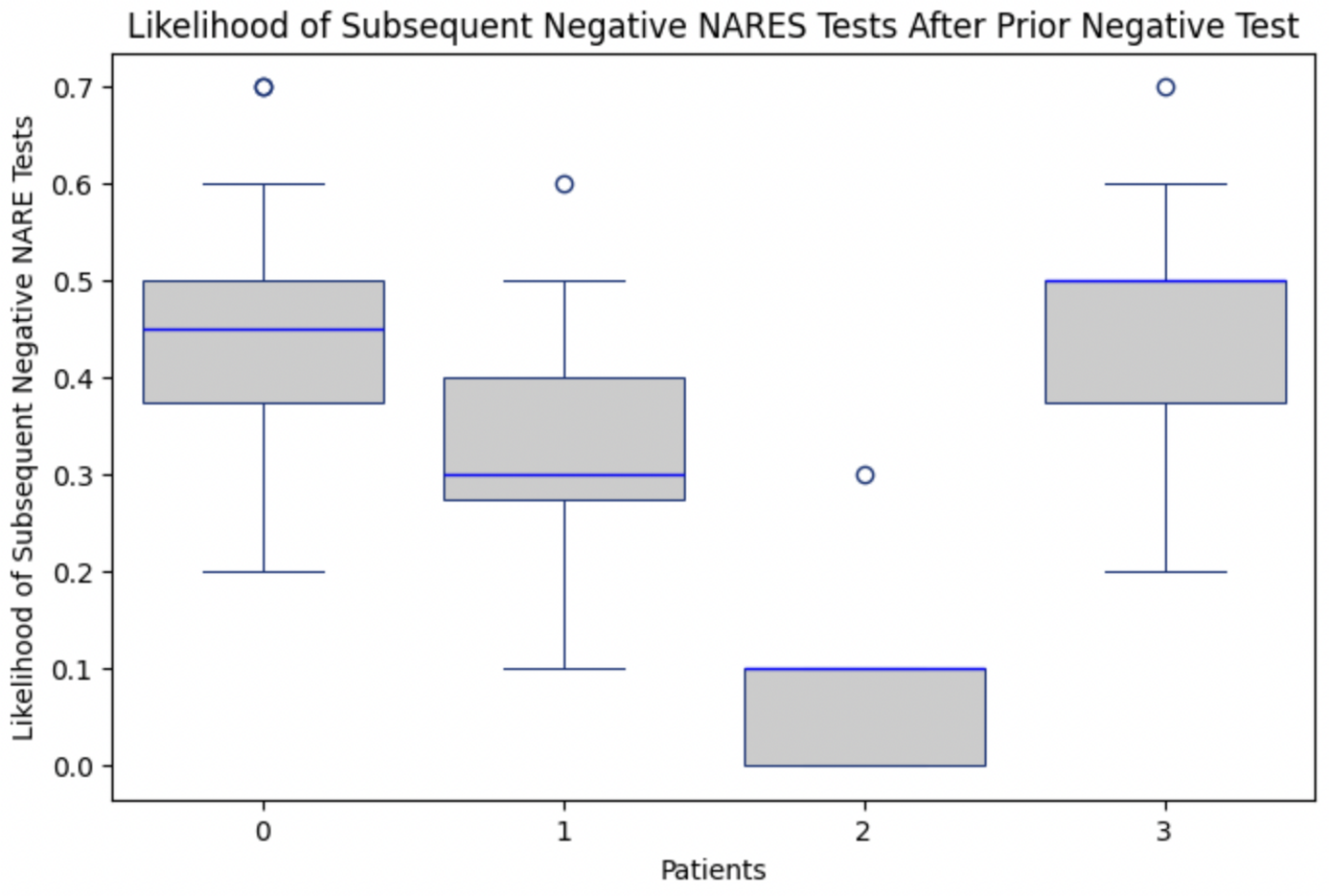}
    \caption{Variation in the likelihood of patients staying
MRSA negative during their remaining hospitalization period for selected patients.}
    \label{fig:cs4}
  \end{subfigure}
\caption{Four clinical decision scenarios supported by \tool{}: 
(a) admission risk, (b) risk with longer stay, 
(c) immediate retest probability, and (d) discontinuing isolation.}

  \label{fig:cs_grid}
  \vspace{-1em}
\end{figure*}



\noindent
\paragraph{Case Study 2: (Future Risk with Extended Stay).} \label{ssec:case-study-2} \textit{A patient has been hospitalized for several weeks, and his past MRSA test result is known. Assuming that this person will be in the hospital for at least X more days, what is the likelihood that they will test positive for MRSA in the next X days?}

\noindent
\textbf{A2:} To efficiently answer this question, we create a variant of \tool{},
that runs the stochastic simulation with the additional test result
as input and runs the simulation forward.
A detailed description of the variant can be found in the Appendix. In addition to using the \textit{$\Tcont$} variable to stop the simulation, the variant algorithm also keeps track of the time passed and stops if the cumulative simulated duration crosses $X$ days. Figure~\ref{fig:cs2} shows how the likelihood changes
with an increase in $X$
for a few selected representative patients.
As can be observed from the figure, there is a general trend
that the likelihood increases with increasing duration of stay in the hospital.

\noindent
\paragraph{Case Study 3 (Immediate Retest).}
\textit{An ICU patient had their last MRSA test X days ago. The infection-control team wonders: if we test today, what’s the chance of a positive result?}


\noindent
\textbf{A3:} The above question is an example of a causal question with interventions.
In particular, it requires that we ignore sampling the inter-test time delay
and use X days as the delay.
Additionally, since the question requires the answer to a single test result,
the one to be done ``right now'' the variant of \tool{}
need only execute an iteration of the test result generation.

Figure~\ref{fig:cs3} shows how the likelihood of the testing positive
changes for a few selected representative patients.
We can observe, that for the selected time period,
for some of the patients, the likelihood of testing positive increase
since the last test result.
However, for a subset of them, the likelihood stays constant,
illustrating the variability captured by the method.

\paragraph{Case Study 4 (Discontinuing Precautions).} \textit{A patient previously MRSA-positive has just returned a negative nasal swab. The physician asks: can we safely discontinue isolation? What is the likelihood that the subsequent NARES test will both be negative and they will not have any positive MRSA?}\\
\noindent
\textbf{A4:} The question posed above is particularly important
from the perspective of hospital-acquired infection risk management.
Patients who test positive for MRSA are generally kept in isolation.
This precautionary measure, while important from a MRSA spread risk management perspective
is however, bad for the isolated patient's mental health
and expensive for the hospital administrators.
The methodology presented in the current study can be used
to evaluate the risk of moving a patient off precautionary isolation.


Figure~\ref{fig:cs4} shows the likelihood distribution
of a patient who tested negative to test positive in the future
for four selected patients.
As expected, heterogeneity can be observed between the different patients.

%% file: conclu.tex
\section{Conclusion}\label{sec:conclu}
Hospital associated infections (HAIs) pose a significant burden for healthcare institutions. Here, we present a new approach, \tool{}, for HAI risk prediction using generative sequence modeling based on a novel modular probabilistic program, guided by clinical domain experts.
All prior methods for HAI risk prediction are based on discriminative ML methods, and are not directly useful in clinical practice.
In contrast, \tool{} provides a powerful framework for supporting questions related to HAIs that arise naturally during complex clinical workflows; further, \tool{} gives better performance than prior methods in many metrics, since it explicitly incorporates clinical workflows.
Our methods are more interpretable than prior approaches, which build ad hoc methods for capturing such workflows, and highlight the utility of probabilistic programming methods for clinical applications.

\paragraph{Path to Deployment.} 
\tool{} uses standard data recorded in EHRs, such as information on admission covariates, antibiotic exposure, ICU stays, and treatments such as dialysis; our results on the UVA and MIMIC-III EHR datasets show that \tool{} can be easily adapted to most EHR systems, in spite of differences in how this information is coded.
\tool{} can be easily integrated with current workflows supported in EHR systems such as EPIC.
Individual patient level risk scores and uncertainty bands at key decision points, including admission screening, ongoing risk assessment, retest prioritization, and de-isolation, can be easily computed and made available to clinicians.
Governance spans infection prevention and ID clinicians, nursing leadership, IT/EHR teams, and oversight bodies (IRB, HIPAA). Its modular structure supports site-specific retraining and extension to other HAIs without altering the overall program.

\noindent
\textit{Validation Plan.} Deployment can follow a safe pathway of silent replay, pilot, and gradual scale-up, with ongoing calibration monitoring to demonstrate real-world impact.

\noindent
\textbf{Limitations and Future Work.} 
\tool{} reflects practices from two US hospitals, but needs to be redesigned for other hospital systems, using their specific workflows and patient populations.
Its modular design improves interpretability but reduces expressiveness by avoiding shared latent variables; future work will relax this constraint to capture richer dependencies. Current inference uses stochastic variational methods, which scale well but only approximate posteriors—alternative approaches (e.g., probabilistic circuits) may improve calibration. Extending \tool{} to evolving hospital policies and multi-institutional datasets will further enhance robustness and clinical applicability.

%% file: related.tex
\section{Additional Related Works}\label{sec:related}

Prior work has developed prediction tools for MRSA and other HAIs using EHR data~\cite{Hartvigsen2018EarlyPO,oh:iche18,Monsalve15,kamruzzaman2024improving,chang:plosone11,dubberke:iche11,na:pone15,modayil:gastroenterology11,fan2022random,liu2019influencing,huynh2022domain,ambavane20232814}. While simple ML with careful feature engineering can capture some risks~\cite{Monsalve15,kamruzzaman2024improving,fan2022random}, such models are limited to narrow classification tasks and cannot support the broad queries clinicians face.  

Our work is also related to sequence modeling for clinical prediction, where recent methods include Crossformer ~\cite{zhang2023crossformer}, TSMixer~\cite{Ekambaram_2023}, and Mamba~\cite{mamba,mamba2}, as well as multi-task EHR models such as TPAMTL~\cite{nguyen2021clinical}. However, these methods underperform on our complex clinical tasks, lacking calibration and reliable uncertainty estimates.  



Probabilistic machine learning and generative models help in addressing many such issues, e.g.,~\cite{chen2021probabilistic,pmlr-v149-urteaga21a,Makar_Guttag_Wiens_2018}. \cite{pmlr-v149-urteaga21a} develop a simple but flexible generative model for the prediction of menstrual cycle length in a mobile health setting. Specifically, their approach involves a hierarchical, Generalized Poisson-based generative model, which explicitly incorporates individual behaviors (e.g., missed tracking), as well as population-level information, and produces calibrated individual predictions. \cite{Makar_Guttag_Wiens_2018} show that a generative probabilistic model performs very well for the problem of predicting the activation of an individual in a networked epidemic process. These applications lead to very interpretable results with good performance, and make it easy to incorporate domain knowledge. However, unlike all prior use of generative models in healthcare settings, the models we consider here are much more complex.

%% file: indiv-models.tex
\section{Algorithm of \tool{}}
Algorithm \ref{alg:full-sim} outlines the framework and the process of generating test sequences.

\begin{algorithm}[t]
\small
\begin{algorithmic}[1]
  \Procedure{SimulateFullSequence}{}
    \State \textbf{Input: $\balpha$}
    \State \textbf{Output: $\{r_i\}, \{t_i\}, \{d_i\}, \{\bbeta_i\}$}
    \State $\bbeta_1[\Tab] \sim \cD_{\bbeta_1[\Tab]}(\balpha)$
    \State $\bbeta_1[\Ticu] \sim \cD_{\bbeta_1[\Ticu]}(\bbeta_1[\Tab], \balpha)$
    \State $\bbeta_1[\Tdia] \sim \cD_{\bbeta_1[\Tdia]}(\bbeta_1[\Ticu], \bbeta_1[\Tab], \balpha)$
    \State $t_1 \sim \cD_{t_1}(\balpha, \bbeta_1)$
    \If{$t_1 = 1$}
      \State $r_1 \sim \cD_{r_1}(\balpha, \bbeta_1)$
    \Else
      \State $r_1 \gets 1$
    \EndIf
    \State $i \gets 2$
    \State $\Tcont \gets \cD_{\Tcont}(r_1, \balpha, \bbeta_1)$
    \While{$\Tcont$}
      \If{$r_{i-1} = 0$}
        \State $d_{i-1} \sim \cD_{d_i|-}(\balpha, \bbeta_{i-1})$
      \Else
        \State $d_{i-1} \sim \cD_{d_i|+}(\balpha, \bbeta_{i-1})$
      \EndIf
      \State $\bbeta_i[\Tab] \sim \cD_{\bbeta_i[\Tab]}(\balpha, \bbeta_{i-1}, r_{i-1}, d_{i-1})$
      \State $\bbeta_i[\Ticu] \sim \cD_{\bbeta_i[\Ticu]}(\bbeta_i[\Tab], \balpha, \bbeta_{i-1}, r_{i-1}, d_{i-1})$
      \State $\bbeta_i[\Tdia] \sim \cD_{\bbeta_i[\Tdia]}(\bbeta_i[\Tdia], \bbeta_i[\Tab], \balpha, \bbeta_{i-1}, r_{i-1}, d_{i-1})$
      \State $t_i \sim \cD_{t_i}(\balpha, \bbeta_i, r_{i-1}, d_{i-1})$
      \If{$t_i = 1$}
        \State $r_i \sim \cD_{r_i}(\balpha, \bbeta_i, r_{i-1}, d_{i-1})$
      \Else
        \State $r_i \gets 1$
      \EndIf
      \State $i \gets i + 1$
      \State $\Tcont \gets \cD_{\Tcont}(r_i, \balpha, \bbeta_i)$
    \EndWhile
  \EndProcedure
\end{algorithmic}
\caption{A probabilistic program
  describing the generative process
  of test sequence data
  for a single hospitalization record.\label{alg:full-sim}}
\end{algorithm}

\subsection{Detailed Description of Algorithm 1}
\label{appendix:alg1-description}

Algorithm \ref{alg:full-sim} is a \textbf{probabilistic program} designed to model the generative process of MRSA test sequences within a patient's hospitalization. It takes a patient's initial features, denoted by $\alpha$, as input and generates a complete sequence of test-related variables: test results ($\{r_i\}$), test types ($\{t_i\}$), inter-test time delays ($\{d_i\}$), and features at the time of each test ($\{\beta_i\}$). The model is structured as a collection of smaller, modular probabilistic sub-programs ($\mathcal{D}_{*}$, each of which can be trained independently using Stochastic Variational Inference (SVI) to maximize the Evidence Lower Bound (ELBO). This modular design, a key feature of the model, enhances interpretability, simplifies debugging, and enables faster, concurrent training. The design choices for each sub-program were made in collaboration with clinical experts to ensure the models accurately reflect the data generation process and have the correct support for their respective random variables.

\paragraph{Modeling Initial Test-Time Features ($\beta_1$)}
The algorithm first focuses on modeling three critical patient features observable at the time of the first MRSA test, collectively referred to as $\beta_1$. These features are crucial for understanding the patient's condition at the start of the testing sequence.

\begin{enumerate}
    \item \textbf{Days on Antibiotics ($\beta_1[ab]$):} This feature, representing the number of days a patient has been on antibiotics in the past 30 days, is modeled using a \textbf{Truncated Negative Binomial distribution} (truncated at 30). This choice is based on domain knowledge that the observations are overdispersed and truncated count values. The distribution of this feature is visualized in Figure~\ref{fig:ab-days-hist}, which shows a log-log plot revealing a power-law-like decay followed by a higher probability at the truncation point of 30 days, likely due to chronic antibiotic users.
    \item \textbf{Days in ICU ($\beta_1[icu]$):} Similarly, the number of days a patient has spent in the Intensive Care Unit within the past 7 days is modeled using a \textbf{Truncated Negative Binomial distribution} (truncated at 7). This reflects the overdispersed truncated nature of the count values, as seen in Figure~\ref{fig:icu-days-hist}. The distribution also exhibits a power-law trend with a higher probability at the truncation point, suggesting a sub-population of severely ill patients who remain in the ICU for the maximum observed period.
    \item \textbf{On Dialysis ($\beta_1[dia]$):} As this is a boolean variable (yes/no), it is modeled using a \textbf{Bernoulli distribution}. This simple model is sufficient to capture the probability of a patient being on dialysis at the time of the test.
\end{enumerate}

Each of these models is conditioned on the patient's initial features ($\alpha$) and, for subsequent features in the sequence, on the output of previously modeled features within the same test.

\paragraph{Modeling Test Type and Result.}

The algorithm accounts for the data's specific structure, where culture test results are only recorded if positive. A \textbf{Bernoulli model} ($\mathcal{D}_{t_1}$) is used to decide the test type: NARE ($t_1=1$) or culture ($t_1=0$). If the test is a culture test, the result ($r_1$) is deterministically set to positive. Otherwise, if it's a NARE test, the result is sampled from another \textbf{Bernoulli model} ($\mathcal{D}_{r_1}$).

\paragraph{Modeling Sequence Continuation and Inter-Test Delays.}

To handle sequences of variable length, the algorithm uses a \textbf{Bernoulli distribution-based continuation model} ($\mathcal{D}_{cont}$). After each test, a sample from this model determines whether the generation process continues or terminates.

The time elapsed since the last MRSA test, a known predictor of future test results, is modeled separately based on the previous test's outcome. The distributions for inter-test times are shown in Figure~\ref{fig:inter-test-time-hist} and are notably different for positive and negative previous results.

\begin{itemize}
    \item \textbf{Previous test was negative ($r_{i-1} = 0$):} The inter-test time delay ($d_i$) is modeled using a \textbf{mixture of three Log-Normal models} ($\mathcal{D}_{d_i|-}$). This choice captures the clinical practice of repeating tests after one day or one week, an effect that creates distinct peaks in the distribution, as seen in Figure~\ref{fig:inter-test-time-hist}. The means of two of these components are set based on this domain knowledge.
    \item \textbf{Previous test was positive ($r_{i-1} = 1$):} For this case, the inter-test time is modeled using a single \textbf{Log-Normal model} ($\mathcal{D}_{d_i|+}$). The distribution for this case is also shown in Figure~\ref{fig:inter-test-time-hist}.
\end{itemize}

The generation of test-time features ($\beta_i$), test types ($t_i$), test results ($r_i$), and inter-test time delays ($d_i$) continues in a loop until the continuation model ($\mathcal{D}_{cont}$) returns zero, at which point the sequence ends. Each step in the loop is conditioned on the patient's initial features ($\alpha$) and the variables generated in the previous step (e.g., $\beta_{i-1}$, $r_{i-1}$, and $d_{i-1}$).

\subsection{GLM like sub-programs used in Algorithm~\ref{alg:full-sim}}

In this section, we shall describe the different sub-programs
are used in Algorithm~\ref{alg:full-sim}.
While some of the models described in this section
are proper Generalized Linear Models or GLMs, others are not.
In particular, we do not assume (for all the models)
that the output is sampled from a distribution
of the exponential family.
For example, samples $\cD_{\bbeta_i[\Tab]}, \cD_{\bbeta_i[\Ticu]}$
are assumed to be distributed from truncated Negative Binomial
distribution which is not in the exponential family.
\subsubsection{Bernoulli models.}
The sub-programs ---
$\cD_{\bbeta_1[\Tdia]}$,
$\cD_{t_1}$,
$\cD_{r_1}$,
$\cD_{\Tcont}$,
$\cD_{\bbeta_i[\Tdia]}$,
$\cD_{t_i}$,
and $\cD_{r_i}$ ---
have the following structure:
\begin{align}
  Y \sim \Bernoulli(p)\\
  p = \rho(\bw^T \cdot \bx + c)\\
  (\bw, c) \sim \bNormal(\bmu_Y, \bSigma_Y)
\end{align}
Here, $\Bernoulli(p)$ is the Bernoulli distribution,
$\bw$ and $c$ are the linear model parameters,
$\rho$ is the logistic link function,
and $\bNormal(\bmu_Y, \bSigma_Y)$ is a multivariate normal distribution
used as a variational approximation for the posterior
distribution over $(\bw, c)$.
\subsubsection{Truncated Negative Binomial models.}
The sub-programs ---
$\cD_{\bbeta_1[\Tab]}$,
$\cD_{\bbeta_1[\Ticu]}$,
$\cD_{\bbeta_i[\Tab]}$,
and $\cD_{\bbeta_i[\Ticu]}$, ---
have the following structure:
\begin{align}
  Y =
    \begin{cases}
      y' &\quad\text{if }y' < t\\
      t  &\quad\text{otherwise}
    \end{cases}\\
  y' \sim \NegBinom(n, p)\\
  p = (1.0 + \alpha \mu)^{-1}\\
  n = \alpha^{-1}\\
  \mu = \exp{(\bw^T \cdot \bx + c)}\\
  (\alpha, \bw, c) \sim \bNormal(\bmu_Y, \bSigma_Y)
\end{align}
Here, $\NegBinom(n, p)$ is the Negative binomial distribution,
$t$ is the truncation threshold,
$\bw$ and $c$ are the linear model parameters,
$\exp$ is the exponential link function,
and $\bNormal(\bmu_Y, \bSigma_Y)$ is a multivariate normal distribution
used as a variational approximation for the posterior
distribution over $(\alpha, \bw, c)$.
$\alpha$ here is the over dispersion parameter,
which ensures that for a given mean $\mu$
the negative binomial distribution has a variance
\mbox{$\sigma^2 = \mu + \alpha\mu^2$}.
\subsubsection{Log-Normal models.}
The sub-program $\cD_{d_i|+}$
has the following structure:
\begin{align}
  Y = \exp(y')\\
  y' \sim \Normal(\mu, \sigma)\\
  \mu = \bw^T \cdot \bx + c\\
  (\sigma, \bw, c) \sim \bNormal(\bmu_Y, \bSigma_Y)
\end{align}

Here, $\Normal(\mu, \sigma)$ is the Normal distribution,
$\bw$ and $c$ are the linear model parameters,
$\exp$ is the exponential function,
$\sigma^2$ is the variance parameter,
and $\bNormal(\bmu_Y, \bSigma_Y)$ is a multivariate normal distribution
used as a variational approximation for the posterior
distribution over $(\sigma, \bw, c)$.

\subsubsection{Mixture of Log-Normal models}
The sub-program $\cD_{d_i|+}$
has the following structure:

\begin{align}
  Y = \exp(y')\\
  y' \sim \Normal(\mu_k, \sigma_k)\\
  k \sim \Multinomial_3(p_1, p_2, p_3)\\
  p_i = \frac{e^{z_i}}{\sum_i e^{z_i}} \quad i\in\{1,2,3\}\\
  z_i = \bw_{z_i}^T \cdot \bx + c_{z_i} \quad i\in\{1,2,3\}\\
  \mu_3 = \bw_{\mu_3}^T \cdot \bx + c_{\mu_3}\\
  (\bw_{z_i}, c_{z_i}, \bw_{\mu_3}, c_{\mu_3}, \sigma_i) \sim \bNormal(\bmu_Y, \bSigma_Y)
\end{align}

Here the index $i\in\{1,2,3\}$ represents the mixture components.
$\bw_{z_i}, c_{z_i}, \bw_{\mu_3}, c_{\mu_3}$ and $\sigma_i$
are the model parameters,
$\Normal(\mu, \sigma)$ is a Normal distribution,
and $\Multinomial_3(p_1, p_2, p_3)$ is a Multinomial distribution
with 3 components.
The parameters $\mu_1$ and $\mu_2$ are known as priori.
$\bNormal(\bmu_Y, \bSigma_Y)$ is a multivariate normal distribution
used as a variational approximation for the posterior
distribution over the model parameters
$(\bw_{z_i}, c_{z_i}, \bw_{\mu_3}, c_{\mu_3}, \sigma_i)$.

%% file: appendix-method.tex
\section{Dataset}
\paragraph{\UVADataset{}.}
The \UVADataset{} has been created from Electronic Health Records
from the hospital attached to the author's institution.
The use and access of the dataset is restricted
by the Institutional Review Board (IRB) at the author's institution,
and may only be used for authorized research studies ---
such as the one presented in the current paper.
The \UVADataset{} covers over 27K patient records,
37K hospitalization records,
and spans over 12 years -- 2012 to 2022.
The data for each hospitalization is divided into two parts:
(a) \emph{on admission data} and (b) \emph{clinical event data}.

\emph{On admission data} includes patient demographics and visit information.
Patient demographics include information about the age, gender, race, ethnicity, etc.
The visit information includes admission reason, admission type, admission source, etc.
This data is used in the model to create admission time features $\balpha$.


\emph{Clinical event data} includes events noted during hospital stays, such as:
procedures (including ICU stays), medication use (including use of antibiotics),
comorbidity identification, and laboratory tests (including MRSA tests).
This data is used in the model for test time features $\bbeta$,
as well as the test types $t_i$ and test results $r_i$.

\emph{MRSA test data} involves two different kind of tests:
a) clinical culture test and b) surveillance test --- including nasal anterior nares test (NARE)
The clinical culture tests present in the dataset only identify MRSA-positive cases.
However, for NARE tests both positive and negative test results are included.

\paragraph{The MIMIC-III v1.4 dataset.}
The MIMIC-III dataset is a large freely-available dataset
consisting of deidentified data from over 40K patients
from Beth Israel Deaconess Medical Center between 2001 and 2012~\cite{johnson2016mimic}.
The dataset has been made public by the authors for clinical and epidemiological research
and contains hospitalization records including demographics, vital signs, and laboratory records.
Although, the dataset is much more limited and different in structure than \UVADataset{}, it contains all features used by the models presented in the current paper. 
Table~\ref{table:patient_numbers} summarizes key statistics of the UVA and MIMIC-III datasets used in our study. It highlights differences in patient count, hospitalization sequences, and MRSA test distributions across the two cohorts.

\begin{table}[t]
\centering
\small
\begin{tabular}{l|r|r}
  \toprule
  & {\UVADataset{}} & { MIMIC-III v1.4} \\
  \midrule
  Year & $2010--2022$ & $2001--2012$\\
  \# Patients & $27,612$ & $46,520$ \\
  \# Hosp.~Sequences & $37,237$ & $58,976$\\
  \# MRSA tests & $42,359$ & $32,167$\\
  \# Positive MRSA tests & $6,946$ & $4,284$\\
  \bottomrule
\end{tabular}
\caption{Overview of the datasets used for modeling hospitalization sequence task.}
\label{table:patient_numbers}
\end{table}

\subsection{Feature Construction}\label{sec:feature_construct}

For the current study we use the following \textbf{admission time features} $\balpha$:
\begin{itemize}
  \item Gender at the birth of the patient
  \item Age of the patient (in years)
  \item Admission type for current hospitalization,
    with the following possible values:
    `emergency' admission, scheduled `elective' procedure, `newborn' child, and `other'.
  \item Whether arriving from a health care facility
  \item Has a history of cerebrovascular disease (such as stroke)
  \item Has diabetes with or without complications
  \item Was in hospital within past 90 days
  \item Was tested MRSA positive in the past 90 days
\end{itemize}

\noindent
Additionally, the following \textbf{test time features} are used $\bbeta$:
\begin{itemize}
  \item Number of days on antibiotics in the past 30 days
  \item Number of days spent in ICU in the past 7 days
  \item Whether on kidney dialysis treatment in the past 7 days
\end{itemize}

\noindent
While the datasets available (specially \UVADataset{}) is much richer,
the above features were considered to limit the complexity of the model
as well as because of the known relevance of the given features
as predictors for MRSA.

\section{Baseline Details}

To understand the effectiveness of the probabilistic program
presented in Algorithm~\ref{alg:full-sim}
we compare it with three state-of-the-art
deep neural network-based generative sequence models.
As discussed in Section~\ref{sec:related},
while many methods have been proposed recently
on generative sequence modeling,
these models were chosen because of:
recency,
availability of usable implementation,
and compatibility with the task at hand.

\textit{Crossformer~\cite{zhang2023crossformer}}
is a deep neural network model for multivariate time series forecasting.
Crossformer uses Informer --- an efficient Transformer alternative ---
for long sequence time-series forecasting.

\textit{TSMixer~\cite{Ekambaram_2023}} is a multivariate time series
forecasting system inspired by MLP-Mixer~\cite{mlpmixer} model from computer vision literature.
Similar to MLP-Mixer, TSMixer utilizes feed-forward neural networks
for the multivariate time series prediction
and has shown significant improvements
over state-of-the-art MLP and Transformer models.

\textit{Mamba~\cite{mamba,mamba2}}
is a recent architecture for generative sequence modeling
and an alternative to the Transformer architecture.
Mamba has demonstrated significant speed-ups (for inference)
and comparable performance with transformer-based models.

\textit{TPAMTL~\cite{nguyen2021clinical}} (Temporal Probabilistic Asymmetric Multi-Task Learning) is a probabilistic framework for healthcare time-series prediction. By enabling dynamic, uncertainty-guided knowledge transfer across tasks and time steps, TPAMTL gives strong performance and interpretability on clinical datasets like MIMIC III~\cite{johnson2016mimic} and Physionet2012~\cite{citi2012physionet}, offering reliable insights for tasks such as infection prediction, heart failure detection, and mortality risk assessment.

For solving the hospital sequence modeling problem using
the above methods we encode
admission time features, test time features,
and test results as a multi-variate time series.
To accommodate for varying sequence lengths
we pad all sequences such that they are of equal length for the first three baselines. TPAMTL can process time series data of varying sequence lengths and make predictions across multiple tasks.

\begin{table}[t]
  \centering
  \small
\resizebox{0.9\linewidth}{!}{%
\begin{tabular}{l|cc|cc}
  \toprule
  & \multicolumn{2}{c|}{\small \UVADataset{}} & \multicolumn{2}{c}{\small MIMIC-III v1.4 dataset} \\
    \midrule
  {\small \tool{}} & {\small NLL} & {\small Perplexity} & {\small NLL} & {\small Perplexity} \\
  \midrule
   $\cD_{cont}$ & 0.415 & 1.514 & 0.581 & 1.788 \\
  $\cD_{\bbeta_1[1]}$ & 1.140 & 3.128 & 0.345 & 1.412 \\
  $\cD_{\bbeta_1[2]}$ & 1.436 & 4.206 & 1.102  & 3.013 \\
  $\cD_{\bbeta_1[3]}$ & 0.135 & 1.145 & 0.018 & 1.019 \\
  $\cD_{t_1}$ & 0.273 & 1.314 & 0.283 & 1.327 \\
  $\cD_{r_1}$ & 0.296 & 1.345 & 0.026 & 1.027 \\
  $\cD_{d_i|+}$ & 1.560 & 4.760 & 2.006 & 7.433\\
  $\cD_{d_i|-}$ & 1.522 & 4.57&  1.136 & 3.114\\
  $\cD_{\bbeta_i[1]}$ & 2.33 & 10.283 & 1.093 &  2.983\\
  $\cD_{\bbeta_i[2]}$ & 0.899 & 2.458 & 1.378 & 3.967 \\
  $\cD_{\bbeta_i[3]}$ & 0.212 & 1.236 & 0.672 & 1.069 \\
  $\cD_{t_i}$ & 0.375 & 1.455 & 0.323 & 1.382 \\
  $\cD_{r_i}$ & 0.195 & 1.215 & 0.064 & 1.066 \\
  \bottomrule
\end{tabular}
}
  \caption{Negative Log-Likelihood (NLL) and Perplexity of held-out sets
  for the individual sub-programs on \UVADataset{} and MIMIC-III dataset.}\label{tab:gen_metrics}
\end{table}

\section{Modeling Conditional Sequence Generation}\label{sec:appendix-method}
Section~\ref{sec:case-study} presented a number of case studies
that required versions of Algorithm~\ref{alg:full-sim}
conditioned on specific observations.
For the current study, we implement these models efficiently
using variants of Algorithm~\ref{alg:full-sim}.
Here we present these variations along with Case Study 3 \& 4.







Q4: For a prior MRSA-positive person who had a negative NARES test,
What is the likelihood that the subsequent NARES test
will both be negative
and they will not have any positive MRSA tests during that hospitalization?


\noindent
A4: The question posed above is particularly important
from the perspective of hospital-acquired infection risk management.
Patients who test positive for MRSA are generally kept under isolation.
This precautionary measure while important from a MRSA spread risk management perspective
is however bad for the isolated patient's mental health
and expensive for the hospital administrators.
The methodology presented in the current study can be used
to evaluate the risk of moving a patient off precautionary isolation.

Figure~\ref{fig:cs4} shows the likelihood distribution
of a patient who tested negative to test positive in the future
for four selected patients.
As expected, heterogeneity can be observed between the different patients.

\begin{algorithm}
\small
\begin{algorithmic}[1]
  \Procedure{SimulatePartialSequenceA}{}
    \State \textbf{Input: $\balpha, \bbeta_1, r_1, \tau_p, \tau_m$}
    \State \textbf{Output: $\{r_i\}, \{t_i\}, \{d_i\}, \{\bbeta_i\}$}
    \State $\Tcont \gets \cD_{\Tcont}(r_1, \balpha, \bbeta_1)$
    \If{$ \text{not } \Tcont$}
        \State \textbf{return}
    \EndIf
    \If{$r_1 = 0$}
       \State $d_1 \sim \cD_{d_i|-}(\balpha, \bbeta_{i-1} \mid d_1 \geq \tau_p)$
    \Else
       \State $d_1 \sim \cD_{d_i|+}(\balpha, \bbeta_{i-1} \mid d_1 \geq \tau_p)$
    \EndIf
    \If{$d_1 > \tau_p + \tau_m$}
        \State \textbf{return}
    \EndIf
    \State $\tau_r \gets \tau_p + \tau_m - d_1$
    \State $i \gets 2$
    \Loop
      \State $\bbeta_i[\Tab] \sim \cD_{\bbeta_i[\Tab]}(\balpha, \bbeta_{i-1}, r_{i-1}, d_{i-1})$
      \State $\bbeta_i[\Ticu] \sim \cD_{\bbeta_i[\Ticu]}(\bbeta_i[\Tab], \balpha, \bbeta_{i-1}, r_{i-1}, d_{i-1})$
      \State $\bbeta_i[\Tdia] \sim \cD_{\bbeta_i[\Tdia]}(\bbeta_i[\Tdia], \bbeta_i[\Tab], \balpha, \bbeta_{i-1}, r_{i-1}, d_{i-1})$
      \State $t_i \sim \cD_{t_i}(\balpha, \bbeta_i, r_{i-1}, d_{i-1})$
      \If{$t_i = 1$}
        \State $r_i \sim \cD_{r_i}(\balpha, \bbeta_i, r_{i-1}, d_{i-1})$
      \Else
        \State $r_i \gets 1$
      \EndIf
      \State $\Tcont \gets \cD_{\Tcont}(r_i, \balpha, \bbeta_i)$
      \If{$ \text{not } \Tcont$}
        \State \textbf{return}
      \EndIf
      \If{$r_i = 0$}
        \State $d_i \sim \cD_{d_i|-}(\balpha, \bbeta_{i-1})$
      \Else
        \State $d_i \sim \cD_{d_i|+}(\balpha, \bbeta_{i-1})$
      \EndIf
      \State $\tau_r \gets \tau_r - d_i$
      \If{$\tau_r < 0$}
        \State \textbf{return}
      \EndIf
      \State $i \gets i + 1$
    \EndLoop
  \EndProcedure
\end{algorithmic}
\caption{A probabilistic program describing the generative process for partial test sequence generation
  for a patient whose previous test result $r_1$,
  done $\tau_p$ time ago, is known.
  Additionally it is assumed that the patient will stay in the hospital
  for $\tau_m$ more time.\label{alg:cond-sim-a}}
\end{algorithm}

\paragraph{Algorithm~\ref{alg:cond-sim-a}.} presents a conditional version of
the probabilistic program Algorithm~\ref{alg:full-sim}
that starts the simulation after the first test result $r_1$ has been
already observed.
It keeps track of the time period simulated ensuring that
the simulation is terminated after $\tau_m$ simulation time
has been exhausted.
This version is used to efficiently answer the question in Section~\ref{ssec:case-study-2}.

\begin{algorithm}
\small
\begin{algorithmic}[1]
  \Procedure{SimulatePartialSequenceB}{}
    \State \textbf{Input: $\balpha, \bbeta_1, r_1, \tau_p$}
    \State \textbf{Output: $\{r_2\}, \{t_2\}, \{d_2\}, \{\bbeta_2\}$}
    \State $d_1 \gets \tau_p$
    \State $\bbeta_2[\Tab] \sim \cD_{\bbeta_i[\Tab]}(\balpha, \bbeta_{1}, r_{1}, d_{1})$
    \State $\bbeta_2[\Ticu] \sim \cD_{\bbeta_i[\Ticu]}(\bbeta_2[\Tab], \balpha, \bbeta_{1}, r_{1}, d_{1})$
    \State $\bbeta_2[\Tdia] \sim \cD_{\bbeta_i[\Tdia]}(\bbeta_2[\Tdia], \bbeta_2[\Tab], \balpha, \bbeta_{1}, r_{1}, d_{1})$
    \State $t_2 \gets 1$
    \State $r_2 \sim \cD_{r_i}(\balpha, \bbeta_2, r_{1}, d_{1})$
  \EndProcedure
\end{algorithmic}
\caption{A probabilistic program describing
  the generative process
  for a patient's next test result
  given his previous test result $r_1$, done $\tau_p$ time ago, is known.
  Additionally, we assume that the next test is done immediately.\label{alg:cond-sim-b}}
\end{algorithm}


\begin{algorithm}
\small
\begin{algorithmic}[1]
  \Procedure{SimulatePartialSequenceC}{}
    \State \textbf{Input: $\balpha, \bbeta_1, r_1=0, \tau_p$}
    \State \textbf{Output: $\{r_i\}, \{t_i\}, \{d_i\}, \{\bbeta_i\}$}
    \State $d_1 \sim \cD_{d_i|-}(\balpha, \bbeta_{i-1} \mid d_1 \geq \tau_p)$
    \State $\bbeta_2[\Tab] \sim \cD_{\bbeta_i[\Tab]}(\balpha, \bbeta_{1}, r_{1}, d_{1})$
    \State $\bbeta_2[\Ticu] \sim \cD_{\bbeta_i[\Ticu]}(\bbeta_2[\Tab], \balpha, \bbeta_{1}, r_{1}, d_{1})$
    \State $\bbeta_2[\Tdia] \sim \cD_{\bbeta_i[\Tdia]}(\bbeta_2[\Tdia], \bbeta_2[\Tab], \balpha, \bbeta_{1}, r_{1}, d_{1})$
    \State $t_2 \sim \cD_{t_i}(\balpha, \bbeta_2, r_{1}, d_{1})$
    \If{$t_2 = 1$}
        \State $r_2 \sim \cD_{r_i}(\balpha, \bbeta_2, r_{1}, d_{1})$
    \Else
        \State $r_2 \gets 1$
    \EndIf
    \State $i \gets 3$
    \State $\Tcont \gets \cD_{\Tcont}(r_2, \balpha, \bbeta_2)$
    \While{$\Tcont$}
      \If{$r_{i-1} = 0$}
        \State $d_{i-1} \sim \cD_{d_i|-}(\balpha, \bbeta_{i-1})$
      \Else
        \State $d_{i-1} \sim \cD_{d_i|+}(\balpha, \bbeta_{i-1})$
      \EndIf
      \State $\bbeta_i[\Tab] \sim \cD_{\bbeta_i[\Tab]}(\balpha, \bbeta_{i-1}, r_{i-1}, d_{i-1})$
      \State $\bbeta_i[\Ticu] \sim \cD_{\bbeta_i[\Ticu]}(\bbeta_i[\Tab], \balpha, \bbeta_{i-1}, r_{i-1}, d_{i-1})$
      \State $\bbeta_i[\Tdia] \sim \cD_{\bbeta_i[\Tdia]}(\bbeta_i[\Tdia], \balpha, \bbeta_{i-1}, r_{i-1}, d_{i-1})$
      \State $t_i \sim \cD_{t_i}(\balpha, \bbeta_i, r_{i-1}, d_{i-1})$
      \If{$t_i = 1$}
        \State $r_i \sim \cD_{r_i}(\balpha, \bbeta_i, r_{i-1}, d_{i-1})$
      \Else
        \State $r_i \gets 1$
      \EndIf
      \State $i \gets i + 1$
      \State $\Tcont \gets \cD_{\Tcont}(r_i, \balpha, \bbeta_i)$
    \EndWhile
  \EndProcedure
\end{algorithmic}
\caption{A probabilistic program
    describing the generative process for partial test sequence generation
    for a patient whose previous test result $r_1$, done $\tau_p$ time ago, is known.
    Additionally, it is assumed that this patient will have at least one more test.\label{alg:cond-sim-c}}
\end{algorithm}



\paragraph{Algorithm~\ref{alg:cond-sim-c}.} presents a conditional version of
the probabilistic program Algorithm~\ref{alg:full-sim}
that is used to efficiently answer the question posed in the case study 4. It assumes that the previous test result is known and is negative. It then simulates the rest of the model forward to generate sequence samples relevant for answering the question.

%% file: limitation.tex
\section{Limitations/Future Work}\label{sec:limitation}
In this section we briefly touch upon
the limitations of the current approach
and point to possible future work to alleviate them.

\paragraph{Domain dependent custom models.}
The models presented in the current study
were developed in collaboration with medical professionals
working on HAI risk management at two US hospitals.
Thus, the models incorporate domain knowledge
about HAI risk management practices at these two institutions.
However, the HAI risk management practices at these hospitals
are likely not representative of most hospitals worldwide,
nor of all hospitals in the US.
Furthermore, these practices may change over time.
Therefore, it is inadvisable to use the presented models, as is,
for modeling HAI risks at different institutions.
Our modeling approach, however, is very flexible,
allowing easy customization for different institutions and use cases.

\paragraph{Independent subprograms.}
The probabilistic programs presented in the current paper
avoid using latent variables at the top level of the program.
This approach makes it easy to build the subprograms,
as they can be trained and validated independently.
However, this also limits the expressiveness of the top-level programs,
which could potentially benefit from sharing latent variables
among the subprograms.
For this study, we made the trade-off in favor of
simplicity, explainability, and training speed.
However, in future work,
we intend to relax this constraint to explore alternate models.

\paragraph{Approximate inference.}
The probabilistic models used in the current paper
are trained using Variational Inference
by optimizing the Evidence Lower Bound (ELBO),
and inference is done by sampling from the posterior.
Approximate inference, via sampling,
is very common in the probabilistic machine learning literature~\cite{variational-inference}.
Recently, probabilistic models that allow for tractable exact inference
have become quite popular~\cite{einsum-networks}.
For example, the `Sum-Product Probabilistic Language' (SPPL)
is a probabilistic programming framework that utilizes these models
for fast exact inference~\cite{sppl}.
The ability of these models to perform exact inference, however,
stems from limiting the model structure to a subset
that is amenable to exact inference.
Thus, in theory, they are not as expressive as general probabilistic models.
In practice, however, they have been shown to be quite flexible.
Thus as future work, we intend to test the suitability of such models for our task.


